\documentclass[letterpaper]{article} 
\usepackage{my}  
\usepackage{times}  
\usepackage{helvet}  
\usepackage{courier}  
\usepackage[hyphens]{url}  
\usepackage{graphicx} 
\usepackage{natbib}  
\usepackage{caption} 
\usepackage{algorithm}
\usepackage{algpseudocode}
\usepackage{amsmath}
\usepackage{amssymb}

\usepackage{caption} 
\usepackage{booktabs} 

\usepackage{newfloat}
\usepackage{listings}
\graphicspath{{pic/}}
\DeclareCaptionStyle{ruled}{labelfont=normalfont,labelsep=colon,strut=off} 
\floatstyle{ruled}
\newfloat{listing}{tb}{lst}{}
\floatname{listing}{Listing}
\nocopyright

\usepackage{cutwin}
\usepackage{enumitem}
\usepackage{amsmath}
\usepackage{booktabs}
\usepackage{multirow}
\usepackage{color}
\usepackage{colortbl}

\usepackage{amsfonts}       
\usepackage{nicefrac}       
\usepackage{microtype}      
\usepackage{xcolor}         
\usepackage{bbding}
\usepackage{pifont}
\usepackage{changepage}

\title{TiKMiX: Take Data Influence into Dynamic Mixture  for  \\ Language Model Pre-training}

\author{
    Yifan Wang,
    Binbin Liu,
    Fengze Liu,
    Yuanfan Guo,
    Jiyao Deng,
    Xuecheng Wu \\
    Weidong Zhou,
    Xiaohuan Zhou\thanks{Corresponding author.},
    Taifeng Wang
}
\affiliations{
    ByteDance \\
    yifanyfwang@gmail.com, zhouxiaohuan@bytedance.com
}

\begin{document}
\maketitle

\begin{abstract}
The data mixture used in the pre-training of a language model is a cornerstone of its final performance. However, a static mixing strategy is suboptimal, as the model's learning preferences for various data domains shift dynamically throughout training. Crucially, observing these evolving preferences in a computationally efficient manner remains a significant challenge. To address this, we propose TiKMiX, a method that dynamically adjusts the data mixture according to the model's evolving preferences. TiKMiX introduces Group Influence, an efficient metric for evaluating the impact of data domains on the model. This metric enables the formulation of the data mixing problem as a search for an optimal, influence-maximizing distribution. We solve this via two approaches: TiKMiX-D for direct optimization, and TiKMiX-M, which uses a regression model to predict a superior mixture. We trained models with different numbers of parameters, on up to 1 trillion tokens. TiKMiX-D exceeds the performance of state-of-the-art methods like REGMIX while using just 20\% of the computational resources. TiKMiX-M leads to an average performance gain of 2\% across 9 downstream benchmarks. Our experiments reveal that a model's data preferences evolve with training progress and scale, and we demonstrate that dynamically adjusting the data mixture based on Group Influence, a direct measure of these preferences, significantly improves performance by mitigating the ``under digestion" of data seen with static ratios.
\end{abstract}

\section{Introduction}

The availability of large-scale public datasets has been a key factor in the creation of Large Language Models (LLMs). The pre-training data for LLMs is predominantly sourced from the internet \cite{wettig2025organize,yu2025craw4llm}, encompassing a wide range of materials such as academic papers \cite{tirumala2023d4}, books \cite{tirumala2023d4}, and more. The mixture ratio of data from these different domains has a significant impact on the capabilities of an LLM \cite{zhang2025survey,liu2025comprehensive,bai2024survey}. For instance, the authors of GPT-3 \cite{floridi2020gpt}  considered Wikipedia to have very high-quality data and consequently decided to increase its proportion in the training dataset. REGMIX \cite{liu2024regmix} uses results from small-scale experiments to automatically set its mixing ratios, but it does not account for dynamic changes in the state of the model \cite{yu2024mates,zhangharnessing}. This leads to a critical research question: How can we dynamically select the training data for a model based on its evolving data preferences in a manner that is both scalable and efficient?

\begin{figure}[t!]
    \centering 
    \hspace*{-0.3cm} 
    \includegraphics[width=0.50\textwidth]{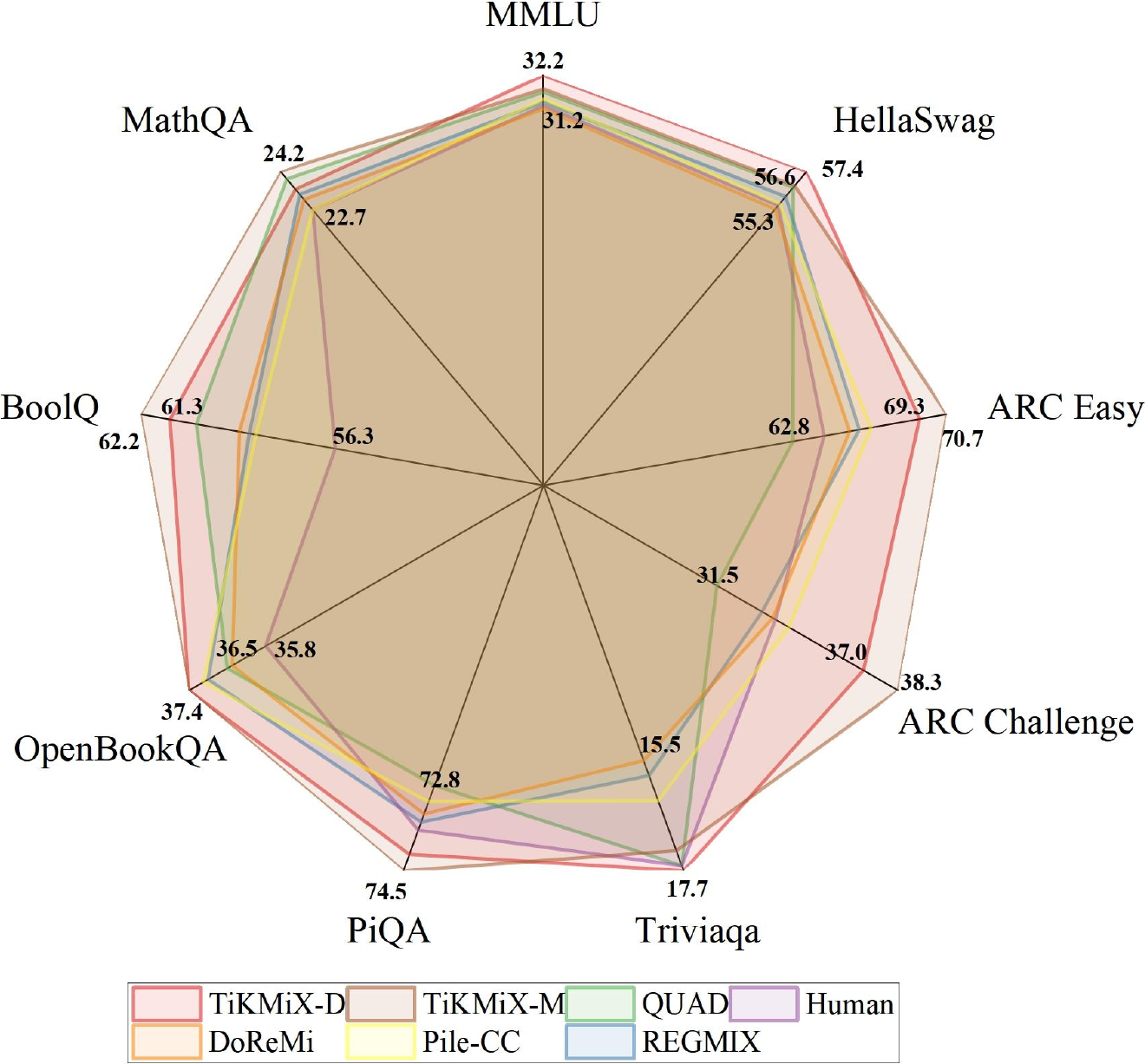} 
    \caption{Performance Comparisons of our TiKMiX versus SOTA Data Mixing Strategies for Pre-training a 1B Parameter Language Model with 1T Tokens.} 
    \label{fig:led} 
\end{figure}

Prior research  \cite{xie2023doremi,fan2023doge,team2024qwen2,albalak2023efficient} has leveraged small proxy models to determine domain weights for large-scale language models. This approach is computationally expensive, as it requires training these proxy models on massive datasets, often exceeding 100 billion tokens. Some methods assume that the relative performance of data mixtures is stable across different model scales and training durations \cite{liu2024regmix}, they neglect the dynamic nature of a model's data preferences as training progresses. Approaches such as ODM \cite{albalak2023efficient} attempt to address this by monitoring training dynamics to guide data allocation, but their iterative nature proves inefficient when dealing with the ever-increasing scale of pre-training data \cite{jin2024survey,wang2025comprehensive}. A significant gap exists in current practices: leading LLMs  \cite{yang2025qwen3,team2025kimi,dubey2024llama} utilize multi-stage pre-training, but they lack a mechanism for rapid, dynamic data re-weighting between stages that aligns with the model's evolving preferences.

We propose a data mixing strategy that dynamically adjusts data proportions during training with minimal computational overhead. To this end, we introduce \textbf{Group Influence}, which efficiently evaluates each domain’s collective impact on validation performance at low computational cost by leveraging gradient accumulation.
This allows us to quantify the model's data preferences at any training stage. 
Building upon this, we introduce \textbf{TiKMiX}, a method that dynamically adjusts the data mixing strategy by framing it as an optimization problem: finding the data combination that maximizes positive influence. 
We devise two approaches to solve this: \textbf{TiKMiX-D}, which directly optimizes a weighted sum of influences from individual domains to find the best mixing ratios; and the more sophisticated \textbf{TiKMiX-M}, which uses TiKMiX-D's output as a starting point, conducts perturbation experiments in its vicinity, and then uses a regression model to fit the relationship between mixing ratios and performance, thereby predicting a globally optimal mixture for subsequent large-scale training.

With the proposed TiKMiX, we can dynamically adjust the data mixture strategy throughout the model's entire pre-training cycle, adapting to changes in model scale and training stage. Following previous work \cite{bai2024multi,kang2024autoscale,diao2025climb,tao2025merge}, we trained models with varying parameter sizes, scaling up to 1 trillion tokens. TiKMiX-D outperforms state-of-the-art methods such as REGMIX while requiring only 20\% of the computational resources. TiKMiX-M achieves an average performance improvement of 2\% across 9 downstream benchmarks as shown in Fig.~\ref{fig:led}. Moreover, we further discuss the feasibility and implications of employing TiKMiX in even larger-scale models.
Our experiments also revealed several key phenomena:(1)A model's data preferences change as training progresses;(2)Models of different scales exhibit different patterns of preference change;(3) Dynamically adjusting the data mixture promotes a more thorough learning of the data by the model. In conclusion, the main contributions of this paper can be summarized as follows:
\begin{itemize}
\item[$\bullet$] We propose \textbf{Group Influence}, a novel and efficient method for observing and quantifying the dynamic preferences of Large Language Models for different data domains during the pre-training process.

\item[$\bullet$] We designed \textbf{TiKMiX}, a dynamic data mixture framework that leverages the observations from Group Influence to adaptively adjust data ratios, aiming to balance the model's performance across multiple tasks.

\item[$\bullet$] Extensive experiments demonstrate that our method not only significantly enhances model performance but also provides profound insights into how a model's data preferences evolve with the training process and model scale, thereby validating the effectiveness of dynamically adjusting data proportions.

\end{itemize}

\label{sec:intro}
\section{Related Work}
\subsection{Influence Function}

Influence Functions offer a mathematically grounded method to estimate the effect of training data on model predictions without costly retraining  \cite{koh2017understanding}. Their application to high-dimensional models like Large Language Models (LLMs) has been hampered by the computational challenge of inverting the Hessian matrix. Recent work has overcome this barrier through scalable approximation techniques. Notably, the work by Anthropic \cite{grosse2023studying} adapted EK-FAC \cite{george2018fast}, an efficient Hessian approximation, to successfully apply influence functions to 50B-parameter Transformer models. This breakthrough established influence functions as a viable tool for performing data attribution at the scale of modern LLMs, enabling the identification of specific pre-training data that drives model outputs \cite{kou2025data, choe2024your,lin2024data}. However, computation at the sample level incurs prohibitive overhead in large-scale pre-training scenarios. Therefore, we propose \textbf{Group Influence}, a method that extends influence functions to groups of data. By leveraging gradient accumulation techniques, Group Influence can efficiently evaluate the collective impact of an entire data domain with relatively low computational cost. This allows us to quantify the model's current data preferences.

\begin{figure*}[!t] 
    \centering
    \includegraphics[width=\textwidth,height=7cm]{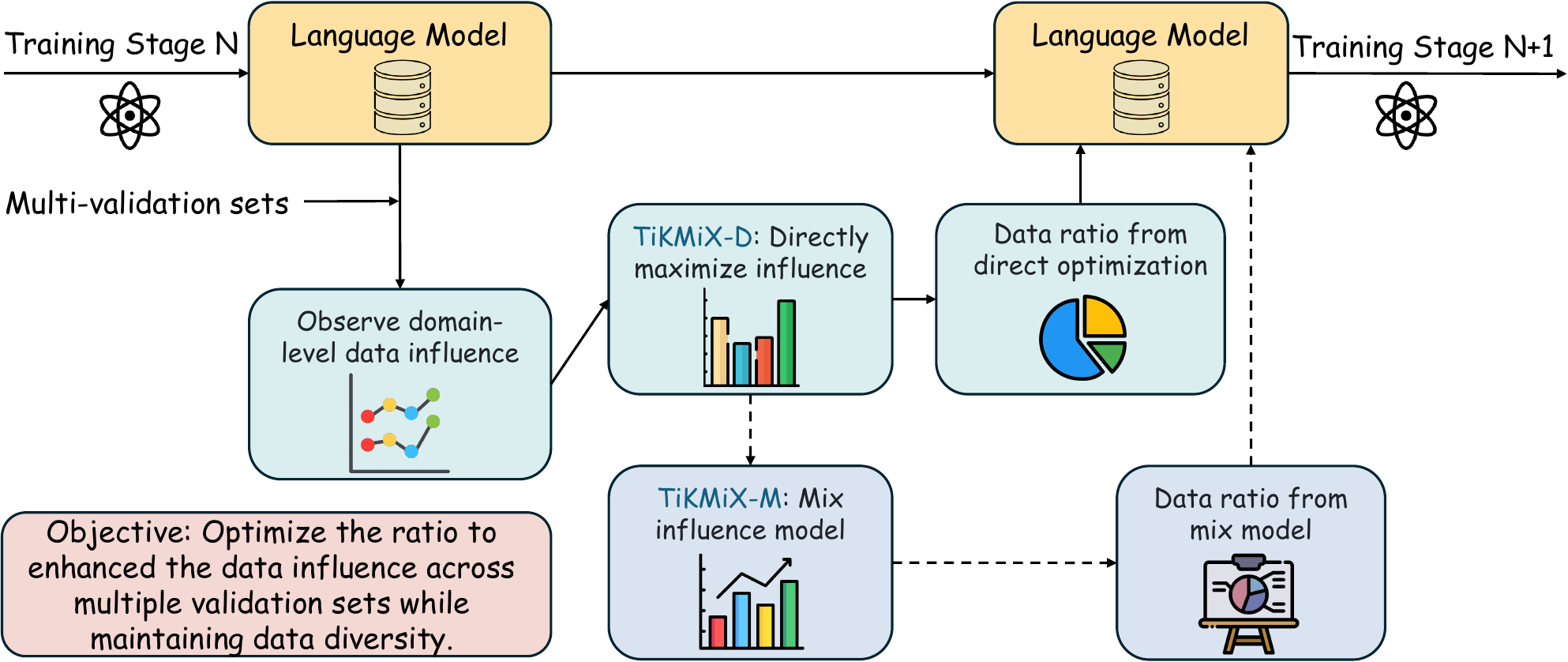}
    \caption{The process involves periodically measuring domain contributions via Group Influence and adjusting the data mixture to maximize learning efficiency.}
    \label{fig:flow}
\end{figure*}

\subsection{Data Selection and Mixing}
\label{subsec:data_selection}

Strategic curation of training data significantly enhances model performance  \cite{koh2017understanding, albalak2023efficient}. For pre-training Large Language Models (LLMs), data curation methods are commonly categorized by granularity:
\textbf{Token-level Selection:} The most fine-grained approach, which filters individual tokens according to specific criteria  \cite{lin2024rho}.
\textbf{Sample-level Selection:}  Methods include heuristic-based approaches  \cite{sharma2024text, soldaini2024dolma} and learning-based techniques employing optimization algorithms  \cite{chen2024take, shao2024balanced}. Additionally, approaches such as MATES  \cite{yu2024mates} utilize model-derived signals (e.g., perplexity or loss) to inform selection  \cite{marion2023less, ankner2024perplexed}.
\textbf{Group-level Selection:} This method partitions data into groups or domains and seeks optimal mixing ratios. Earlier work relied on manually defined ratios, while recent advances favor learning-based strategies. Offline methods like REGMIX  \cite{liu2024regmix} and DoReMi  \cite{xie2023doremi} use proxy models to assign static group weights, whereas dynamic methods such as Quad  \cite{zhangharnessing} and ODM  \cite{albalak2023efficient} iteratively adjust weights during training.
Current mainstream pre-training pipelines are typically divided into multiple stages but often lack a mechanism to dynamically adjust the data mixture ratio based on the model's state in different stages. Our proposed method, TiKMiX, is a semi-offline, group-level selection approach that dynamically adjusts the data mixture ratio across multiple training stages. Unlike fully dynamic methods that require repeated iterative updates, TiKMiX directly optimizes the mixture ratio based on the model’s current data preferences, enabling efficient adaptation without multiple rounds of adjustment.

\label{sec:Related Work}

\section{Methodology}

In this section, we introduce TiKMiX, a framework for dynamically optimizing the data mixture during large language model pre-training as shown in Fig.~\ref{fig:flow}. Our approach is centered on a novel metric, Group Influence, designed to efficiently measure the real-time contribution of each data domain to the model's learning. We formulate the dynamic data mixture problem as an optimization task aimed at maximizing this Group Influence. To solve this, we propose two distinct methods : TiKMiX-D, which directly optimizes the mixture based on influence scores, and TiKMiX-M, which leverages a regression model for a computationally efficient approximation. We first define the problem setup and Group Influence, then elaborate on these two optimization strategies.

\label{sec:methodology}

\begin{figure*}[!t] 
    \centering
    \includegraphics[width=\textwidth]{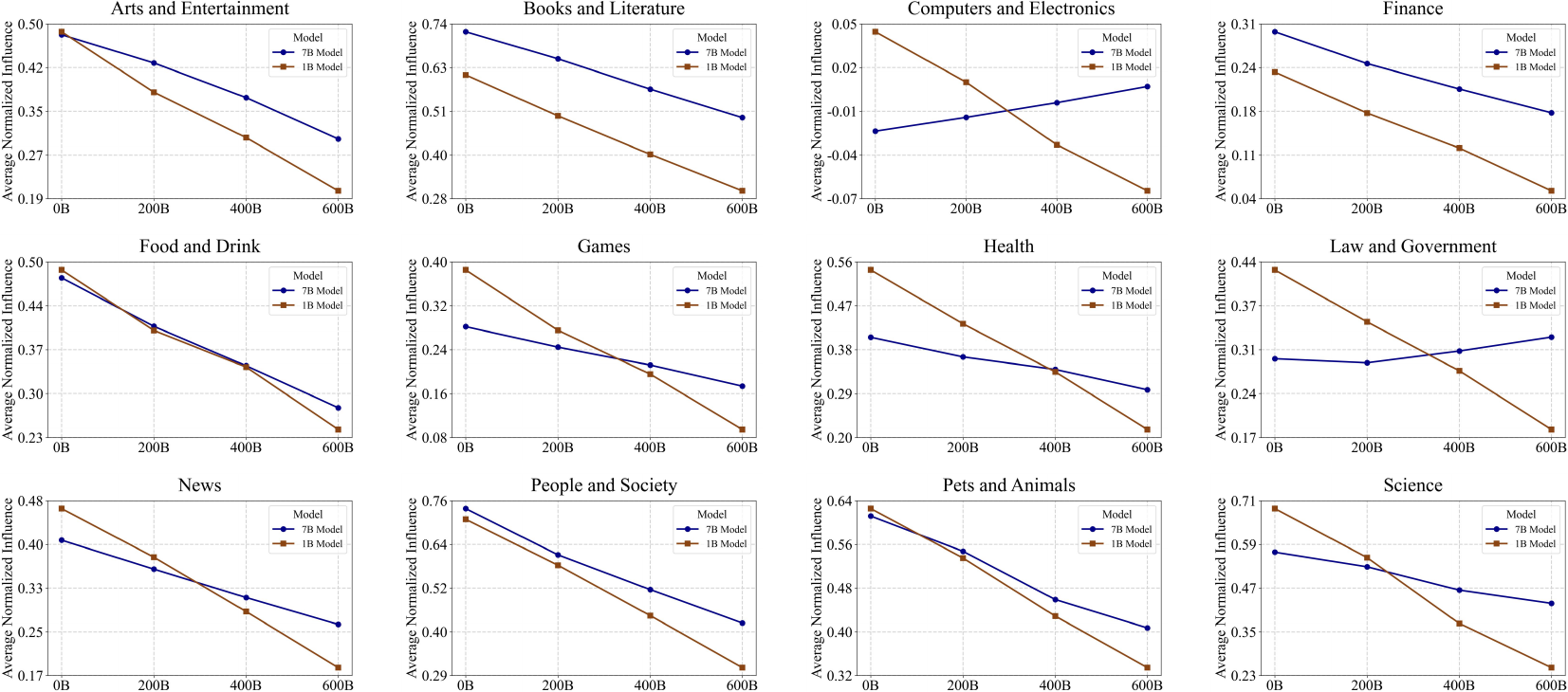}
    \caption{The impact of different pre-training data domains on the validation set as training progresses.}
    \label{fig:case}
\end{figure*}

\subsection{Group Influence}

Group Influence function extends the classical influence function framework from individual data points to cohesive groups of data. We first establish the theoretical motivation for this extension, then provide a rigorous mathematical derivation of Group Influence, and finally, discuss its computational properties.

Influence functions offer a principled and computationally efficient method for estimating the effect of a single training instance on a model's parameters or predictions  \cite{koh2017understanding}. By approximating the change in model parameters resulting from upweighting a training point $z$, they provide valuable insights into model behavior without the need for retraining. However, many complex model behaviors, such as systemic bias, factual recall, or vulnerability to specific adversarial attacks, are not attributable to a single, isolated training example. Instead, they often emerge from the collective effect of a \textit{group} of semantically related instances. A linear summation of individual influence scores, i.e., $\sum_{z_i \in S} I(z_i)$, is insufficient as it fails to capture the non-trivial interactions between data points during optimization. The collective gradient of a group can shape the loss landscape in a manner distinct from the sum of its constituent parts.
To quantify the consolidated impact of a data subset $S$ as a single entity, we define the Group Influence function. Let a model, parameterized by $\theta \in \mathbb{R}^d$, be trained on a dataset $D = \{z_1, \dots, z_N\}$ by minimizing an empirical risk objective $J(\theta)$:
\begin{equation}
    \theta^* = \arg\min_{\theta} J(\theta) = \arg\min_{\theta} \frac{1}{N} \sum_{i=1}^{N} \mathcal{L}(z_i, \theta),
    \label{eq:erm}
\end{equation}
where $\mathcal{L}(z_i, \theta)$ is the loss function for instance $z_i$. To measure the influence of a subset $S \subseteq D$, we introduce a perturbed objective where every member of $S$ is simultaneously upweighted by an infinitesimal positive value $\epsilon$. The new optimal parameters $\theta^*_\epsilon$ are found by minimizing this perturbed objective:
\begin{equation}
    \theta^*_\epsilon = \arg\min_{\theta} \left( \frac{1}{N} \sum_{i=1}^{N} \mathcal{L}(z_i, \theta) + \epsilon \sum_{z_j \in S} \mathcal{L}(z_j, \theta) \right).
    \label{eq:perturbed_objective}
\end{equation}
This formulation models a scenario where the training process is nudged to place greater emphasis on the group $S$. For $\epsilon=0$, we recover the original optimal parameters, $\theta^*_{\epsilon=0} = \theta^*$. The influence of group $S$ on the model parameters is then defined as the rate of change of $\theta^*_\epsilon$ with respect to $\epsilon$, evaluated at $\epsilon=0$.
A closed-form expression for this quantity can be derived using the implicit function theorem. The first-order optimality condition for any $\epsilon$ requires that the gradient of the perturbed objective at its minimum $\theta^*_\epsilon$ is zero, which can be formulated as:

\begin{equation}
    \nabla_\theta J_\epsilon(\theta^*_\epsilon, S) = \frac{1}{N} \sum_{i=1}^{N} \nabla_\theta \mathcal{L}(z_i, \theta^*_\epsilon) + \epsilon \sum_{z_j \in S} \nabla_\theta \mathcal{L}(z_j, \theta^*_\epsilon) = 0.
    \label{eq:optimality_condition}
\end{equation}

Differentiating this entire equation with respect to $\epsilon$ via the chain rule yields:
\begin{equation}
    \frac{d}{d\epsilon} \left[ \nabla_\theta J_\epsilon(\theta^*_\epsilon, S) \right] = \nabla^2_\theta J_\epsilon(\theta^*_\epsilon, S) \frac{d\theta^*_\epsilon}{d\epsilon} + \frac{\partial}{\partial\epsilon} \left( \nabla_\theta J_\epsilon(\theta^*_\epsilon, S) \right) = 0.
    \label{eq:total_derivative}
\end{equation}

Evaluating this expression at $\epsilon=0$ (where $\theta^*_{\epsilon=0} = \theta^*$), the Hessian $\nabla^2_\theta J_\epsilon(\theta^*_\epsilon, S)$ simplifies to the Hessian of the original objective, $H_{\theta^*} \triangleq \nabla^2_\theta J(\theta^*)$. The partial derivative term becomes $\sum_{z_j \in S} \nabla_\theta \mathcal{L}(z_j, \theta^*)$. Substituting these into Equation~\ref{eq:total_derivative} gives:
\begin{equation}
    H_{\theta^*} \left. \frac{d\theta^*_\epsilon}{d\epsilon} \right|_{\epsilon=0} + \sum_{z_j \in S} \nabla_\theta \mathcal{L}(z_j, \theta^*) = 0.
    \label{eq:hessian_equation}
\end{equation}
Assuming the Hessian $H_{\theta^*}$ is positive definite and thus invertible, we can solve for the influence of group $S$ on the model parameters:
\begin{equation}
    I_{\text{param}}(S) \triangleq \left. \frac{d\theta^*_\epsilon}{d\epsilon} \right|_{\epsilon=0} = -H_{\theta^*}^{-1} \left( \sum_{z_j \in S} \nabla_\theta \mathcal{L}(z_j, \theta^*) \right).
    \label{eq:param_influence}
\end{equation}
A common practical application is to measure the influence of $S$ on a scalar-valued function of the parameters, $f(\theta)$, such as the loss on a test sample, $f(\theta) = \mathcal{L}(z_{\text{test}}, \theta)$. By applying the chain rule, the influence of $S$ on $f$ is given by:
\begin{equation}
    I_f(S) \triangleq \left. \frac{df(\theta^*_\epsilon)}{d\epsilon} \right|_{\epsilon=0} = \nabla_\theta f(\theta^*)^T \left. \frac{d\theta^*_\epsilon}{d\epsilon} \right|_{\epsilon=0}.
    \label{eq:scalar_chain_rule}
\end{equation}
Substituting Equation~\ref{eq:param_influence} into Equation~\ref{eq:scalar_chain_rule} yields the final expression for the \textbf{Group Influence function}:
\begin{equation}
    I_f(S) = -\nabla_\theta f(\theta^*)^T H_{\theta^*}^{-1} \left( \sum_{z_j \in S} \nabla_\theta \mathcal{L}(z_j, \theta^*) \right).
    \label{eq:group_influence_final}
\end{equation}
The scalar value $I_f(S)$ quantifies the extent to which upweighting the group $S$ during training would increase ($I_f(S) > 0$) or decrease ($I_f(S) < 0$) the value of the function $f$.
A significant computational advantage of Equation~\ref{eq:group_influence_final} is its structure. The term $\sum_{z_j \in S} \nabla_\theta \mathcal{L}(z_j, \theta^*)$ is the \textit{accumulated gradient} of the group $S$. This allows for an efficient implementation where the gradients for all samples within the group are first computed and aggregated. Subsequently, the computationally intensive Hessian-inverse-vector product is performed only once. This structure ensures the computation of Group Influence is scalable, as its cost is not dominated by the cardinality of the group $|S|$.

\subsection{TiKMiX-D: Directly maximize influence}
\label{sec:tikmix-d}
Building upon the Group Influence metric, which quantifies the impact of data domains on the model's state as shown in Fig.~\ref{fig:case}, our objective is to determine an optimal data mixture, represented by a weight vector $w$, that maximizes the aggregate benefits of this influence. To this end, we introduce TiKMiX-D, a method that formulates this task as a multi-objective optimization problem. This approach dynamically adjusts the data mixture for the subsequent training stage to achieve balanced performance improvement, maximize overall gains, and maintain data diversity.
The Group Influence scores, as computed in the previous section, are first organized into an $n \times m$ Influence Matrix, $S$, where $n$ is the number of validation tasks and $m$ is the number of data domains. Each element $S_{ij}$ represents the influence of data domain $d_j$ on validation task $v_i$. Given a data mixture weight vector $w = [w_1, w_2, \dots, w_m]^T$, the expected total influence on each validation task is captured by the vector $P = S \cdot w$. To facilitate fair comparison across tasks of varying scales, we normalize this influence vector. The normalized influence $\hat{P}_i$ for task $v_i$ can be computed as:
\begin{equation}
    \hat{P}_i = \frac{P_i}{\max_{j} S_{ij} + \epsilon},
\end{equation}
where $\epsilon$ is a small constant (e.g., $10^{-8}$) to prevent division by zero.
The optimization objective for TiKMiX-D is formulated as a unified function $L(w)$ that integrates three distinct goals:
\begin{equation}
    L(w) = \alpha \cdot \text{std}(\hat{P}) - \beta \cdot \sum_{i=1}^{n} \hat{P}_i - \gamma \cdot H(w).
\end{equation}
This function balances three key components. First, \textbf{influence uniformity}, measured by the standard deviation of the normalized influence, $\text{std}(\hat{P})$, encourages balanced capability gains across all tasks. Second, \textbf{overall influence gain}, represented by the total sum of normalized influences, $\sum \hat{P}_i$, aims to maximize the model's aggregate performance improvement; hence, its negative is minimized. Third, \textbf{data diversity}, measured by the information entropy of the weight vector, $H(w) = -\sum_{j=1}^{m} w_j \log(w_j)$, promotes a more uniform weight distribution to ensure robust generalization. The hyperparameters $\alpha, \beta,$ and $\gamma$ control the trade-offs among these objectives; in our experiments, they are set to 1 to assign equal importance.
The complete optimization problem is subject to several constraints to ensure a valid and beneficial solution. The weights must be non-negative ($w_j \ge 0$) and sum to one ($\sum w_j = 1$). Furthermore, to guarantee continuous improvement, we enforce a Pareto improvement constraint, ensuring that the influence generated by the new mixture $w$ is no less than that of the prior mixture $w_{\text{prior}}$ for any task, i.e., $S \cdot w \ge S \cdot w_{\text{prior}}$. This leads to the final constrained non-linear optimization problem:
\begin{equation}
\label{eq:tikmixd_optimization}
\begin{aligned}
& \underset{w}{\text{minimize}}
& & \alpha \cdot \text{std}(\hat{P}) - \beta \cdot \sum_{i=1}^{n} \hat{P}_i - \gamma \cdot H(w) \\
& \text{subject to}
& & \sum_{j=1}^{m} w_j = 1 \\
& & & w_j \ge 0, \quad \forall j \in \{1, \dots, m\} \\
& & & S \cdot w \ge S \cdot w_{\text{prior}}. 
\end{aligned}
\end{equation}
We employ the Sequential Least Squares Quadratic Programming algorithm \cite{gupta2018ensemble} to solve this problem, initializing the weights with a uniform distribution. The resulting optimal vector, $w_{\text{best}}$, serves as the dynamic data mixture for the subsequent training stage.

\subsection{TiKMiX-M: Mix influence model}
\label{sec:tikmix-m}

While TiKMiX-D provides an efficient strategy for data mixing through direct optimization, it operates on the assumption that the influences of data domains are linearly additive. This simplification may overlook the mix of different domain, non-linear cross-domain interactions that arise when different data sources are combined. We introduce TiKMiX-M, optimize mixture proportions by modeling the interactions within domain mixtures To more accurately capture these mixture effects.

To explore the model's performance across a diverse range of domain weightings, we generate a set of $N$ candidate mixture vectors. Our approach is anchored by an empirically determined prior weight vector, $w_{\text{orig}} \in \mathbb{R}^D$, where $D$ is the number of domains. For each domain $i$, we define a plausible sampling interval by scaling the original weight. We employ Latin Hypercube Sampling \cite{loh1996latin} within this $D$-dimensional hyperrectangle to efficiently generate candidate vectors, ensuring a uniform and non-collapsing coverage of the parameter space.

Each candidate vector $w_{\text{cand}}$ produced by Latin Hypercube Sampling is subsequently normalized to satisfy the sum-to-one constraint ($\sum_{i=1}^D w_i = 1$), yielding a normalized vector $w_{\text{norm}} = w_{\text{cand}} / \sum_{j=1}^D w_{\text{cand}, j}$. However, this normalization can shift components outside their predefined intervals. Therefore, we implement a rejection sampling scheme, where a normalized vector $w_{\text{norm}}$ is accepted into our final set only if it satisfies the boundary constraints for all dimensions, i.e., $w_{\text{norm}, i} \in [l_i, h_i]$ for all $i \in \{1, \dots, D\}$. This iterative process is repeated until $N$ valid weight vectors that meet both the summation and boundary conditions have been collected, resulting in a robust and well-distributed set of weights for subsequent analysis.

For each generated candidate mixture $w_i$, we calculate its true aggregate influence score, $y_i$, across all validation sets using the Group Influence evaluation method. This score can be the sum of normalized influences, $\sum \hat{P}_i$, or another comprehensive metric.

Following these steps, we obtain a training set $D_{\text{train}} = \{(w_i, y_i)\}_{i=1}^N$. We select LightGBM \cite{ke2017lightgbm}, an efficient gradient boosting decision tree model, as our regression surrogate. This model, $f_{\text{LGBM}}$, is trained to predict the aggregate influence $y$ for given data mixture $w$, i.e., $y = f_{\text{LGBM}}(w)$.We leverage it to efficiently explore the mixture space without performing expensive, true influence evaluations. We design an iterative search algorithm that balances exploration and exploitation to find the optimal mixture.

\begin{algorithm}[h]
\caption{Iterative Search via TiKMiX-M}
\label{alg:tikmix-m-simplified}
\begin{algorithmic}
\State \textbf{Input:} Surrogate $f_{\text{sur}}$, initial mix $w^{(0)}$, iters $T$, samples $N$, exploration $[\alpha_{\min}, \alpha_{\max}]$, top-$k$.
\State \textbf{Output:} Optimized mixture $w^*$.
\State $w_{\text{best}} \leftarrow w^{(0)}$
\State Generate exploration strengths $\{\alpha_t\}_{t=1}^T$ logarithmically from $\alpha_{\max}$ to $\alpha_{\min}$.
\For{$t = 1$ to $T$}
    \State Sample candidates $\{w_i\}_{i=1}^N \sim \text{Dirichlet}(\alpha_t \cdot w_{\text{best}})$.
    \State Predict scores $y_i = f_{\text{sur}}(w_i)$ for each $w_i$.
    \State Let $I_{\text{top-k}}$ be the indices of the top $k$ scores.
    \State $w_{\text{best}} \leftarrow \frac{1}{k} \sum_{i \in I_{\text{top-k}}} w_i$.
\EndFor
\State \textbf{return} $w_{\text{best}}$
\end{algorithmic}
\end{algorithm}

The process is detailed in Algorithm~\ref{alg:tikmix-m-simplified}. We start from the ratio from TiKMiX-D, $w_{\text{best-D}}$. At each step, we sample candidate mixtures on the current best solution. The distribution's concentration parameter is annealed over step, beginning with a large value to encourage global exploration and gradually decreasing to promote local exploitation near the optimum. We employ the surrogate model to evaluate all sampled candidates. The center for the next iteration is then updated to be the average of the top-k candidates with the highest predicted scores. This procedure is repeated until convergence or a maximum number of iterations is reached.

TiKMiX-M not only accounts for non-linear cross-domain interactions but also significantly enhances search efficiency through the surrogate model, enabling it to discover superior solutions within the vast mixture space.

\label{sec:Mthods}
\section{Experiments}
\label{sec:exprs}

This section presents a comprehensive set of experiments designed to validate the effectiveness of our TiKMiX framework. We first outline the experimental setup, including evaluation benchmarks, datasets, and baseline methods. Subsequently, we demonstrate that: (1) the pre-training data mixture significantly impacts downstream task performance; (2) our proposed Group Influence is an effective predictor of downstream performance; and (3) the TiKMiX framework, particularly TiKMiX-D and TiKMiX-M, markedly improves model performance and surpasses existing SOTA methods.

\subsection{Experimental Setup}

\paragraph{Datasets and Models}
Optimizing the data mixture of web-scale corpora is a crucial and highly impactful step in pre-training performant Large Language Models (LLMs). While the diversity of web data presents unique challenges, effective mixing strategies can unlock significant performance gains. To systematically investigate this, we conduct our experiments on the RefinedWeb dataset~\cite{penedo2023refinedweb}, which comprises 26 distinct data domains. Our models, ranging in size from 1B to 7B parameters, are trained on up to 1 trillion tokens. The training process is divided into two distinct stages, each consisting of 500 billion tokens, with a strategic adjustment of the data mixture ratio at the transition point between stages. 
We compare TiKMiX against several representative data mixing strategies:
\textbf{Pile-CC \cite{gao2020pile}}: The original data mixture proposed by the authors of The Pile based on heuristics.
\textbf{REGMIX \cite{liu2024regmix}}: SOTA method that uses a regression model to predict and optimize validation loss for determining the mixture.
\textbf{DoReMi \cite{xie2023doremi}}: a classic dynamic data mixing method that relies on a proxy model. \textbf{QUAD \cite{zhangharnessing}}: a method for dynamic selection during training after clustering data We use the best-reported mixture from their paper, re-normalized to the domains available in our setup.

\paragraph{Downstream Task Evaluation}
To comprehensively evaluate our proposed method, we curated a diverse set of 9 widely recognized downstream benchmarks, which were strategically divided into two categories: in-domain and out-of-domain. This division allows for a rigorous assessment of both the model's core capabilities and its generalization prowess. Our \textbf{in-domain} evaluation suite was designed to cover a wide spectrum of reasoning and knowledge-based tasks. It includes \textbf{MMLU (Massive Multitask Language Understanding)}~ \cite{hendrycks2020measuring}, a challenging benchmark measuring knowledge across 57 diverse subjects; \textbf{HellaSwag}~ \cite{zellers2019hellaswag}, a commonsense reasoning task that involves choosing the most plausible continuation for a given context; \textbf{ARC}~ \cite{clark2018think}, which we evaluate on both the Easy (\textbf{ARC-E}) and the more difficult Challenge (\textbf{ARC-C}) sets of grade-school science questions; and \textbf{TriviaQA}~ \cite{joshi2017triviaqa}, a reading comprehension benchmark requiring models to locate answers within lengthy documents. To evaluate the generalization capabilities of our method, we selected a set of out-of-domain benchmarks. These include \textbf{PiQA}~ \cite{bisk2020piqa}, a commonsense benchmark focused on physical interactions; \textbf{OpenBookQA}~ \cite{mihaylov2018can}, a question-answering task requiring reasoning over a given set of science facts; \textbf{BoolQ}~ \cite{clark2019boolq}, a dataset of naturally occurring yes/no questions; and \textbf{MathQA}~ \cite{amini2019mathqa}, a mathematical reasoning benchmark with multi-step word problems.



\subsection{Group Influence as an Effective Predictor of Performance}
\begin{figure}[t!]
\centering
\includegraphics[width=0.95\linewidth,height=7cm]{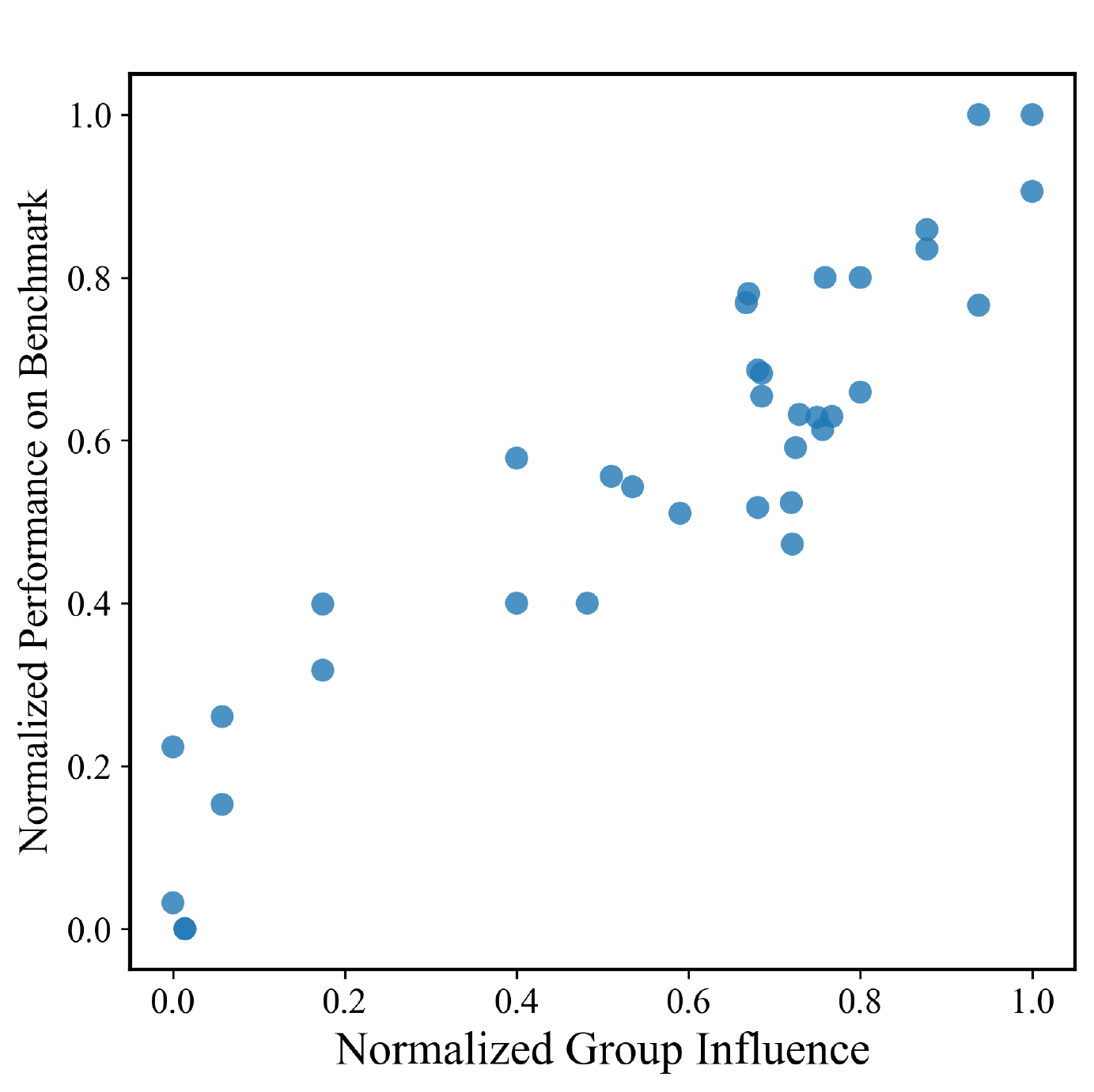}
\caption{Analysis of the Group Influence and actual performance on the benchmark.}
\label{fig:group_benchmark_analysis}
\end{figure}

The core hypothesis of our introduced TiKMiX framework is that maximizing Group Influence can effectively enhance overall downstream task performance. To validate this hypothesis, we calculated the impact of 10 different data mixtures on various benchmarks. As validation, we trained a 1B-parameter model on 500B data using the corresponding mixtures. The normalized scores are shown in Fig.~\ref{fig:group_benchmark_analysis}.
We can clearly observe a strong positive correlation (\textit{i.e.}, Pearson correlation coefficient $\rho =0.789 $) between the total Group Influence and the average downstream scores. This indicates that mixtures generating higher total influence almost invariably lead to better downstream performance. This finding not only confirms the validity of Group Influence as an optimization target but also provides a solid theoretical foundation for the design of our TiKMiX-D and TiKMiX-M.

\begin{table*}[t!]
\centering

\resizebox{\textwidth}{!}{
\begin{tabular}{
    l
    *{8}{c}
}
\toprule
\textbf{Benchmark} & \textbf{Human} & \textbf{DoReMi} & \textbf{Average} & \textbf{QUAD} & \textbf{Pile-CC} & \textbf{REGMiX} & \textbf{TiKMiX-D} & \textbf{TiKMiX-M}\\
\midrule
\multicolumn{9}{l}{\textit{\textbf{In-Domain Benchmarks}}} \\

MMLU  \cite{hendrycks2020measuring} & 31.3 & 31.2 & 30.9 & 31.7 & 31.2 & 31.5 & \textbf{32.2} & 31.8\\
HellaSwag  \cite{zellers2019hellaswag} & 55.5 & 55.3 & 55.9 & 56.5 & 55.6 & 56.0 & \textbf{57.4} & 56.6\\
ARC Easy  \cite{clark2018think} & 64.4 & 65.7 & 64.1 & 62.8 & 63.2 & 66.2 & 69.3 & \textbf{70.7}\\
ARC Challenge  \cite{clark2018think} & 33.7 & 33.6 & 32.1 & 33.5 & 32.7 & 33.2 & 37.0 & \textbf{38.3}\\
Triviaqa  \cite{joshi2017triviaqa} & 17.6 & 15.5 & 17.3 & 17.6 & 16.3 & 15.8 & \textbf{17.7} & 17.3 \\
\hline

\multicolumn{9}{l}{\textit{\textbf{Out-of-Domain Benchmarks}}} \\
PiQA  \cite{bisk2020piqa} & 73.5 & 73.1 & 71.5 & 72.4 & 69.2 & 73.3 & 74.1 & \textbf{74.5}\\
OpenBookQA  \cite{mihaylov2018can} & 35.8 & 36.5 & 34.6 & 36.6 & 37.1 & 37.0 & \textbf{37.4} & \textbf{37.4}\\
Boolq  \cite{clark2019boolq} & 56.3 & 59.2 & 58.3 & 60.5 & 58.7 & 58.9 & 61.3 & \textbf{62.2}\\
MathQA  \cite{amini2019mathqa} & 22.7 & 23.1 & 23.7 & 23.9 & 22.5 & 23.3 & 23.5 & \textbf{24.2}\\
\midrule

Estimated FLOPs & 0 & 4.2e19 & 0 & 2.3e18 & 0 & 3.7e18 & 7.2e17 & 3.2e18 \\
Average Perf. & 43.4 & 43.7 & 43.2 & 43.9 & 42.9 & 43.9 & 45.5 & 45.9 \\
Best On & 0/9 & 0/9 & 0/9 & 0/9 & 0/9 & 0/9 & 4/9 & 6/9 \\
\bottomrule
\end{tabular}
}
\caption{Comparison of 1B Parameter Models Trained on 1T Tokens Across Various Benchmarks. The best-performing model on each benchmark is highlighted in bold.}
\label{tab:benchmark-comparison}
\end{table*}

\subsection{TiKMiX Improves Downstream Performance}
Building on the preceding findings, we formally evaluate the two implementations of our TiKMiX framework: \textbf{TiKMiX-D} and \textbf{TiKMiX-M}.
During a 1T-token pre-training process, we dynamically adjusted the data mixture every 200B tokens using TiKMiX. As shown in Table~\ref{tab:benchmark-comparison}, both of our methods significantly outperform all baselines. On average, across 9 benchmarks, \textbf{TiKMiX-D} and \textbf{TiKMiX-M} improved performance by \textbf{1.6\%} and \textbf{2.0\%}, respectively, over the strongest baseline, \textbf{REGMIX}. Notably, on challenging tasks like \textbf{ARC Easy} and \textbf{ARC Challenge}, TiKMiX-M achieved a performance advantage of over 4.8\%.
\subsection{Analysis of Computational Efficiency}
TiKMiX also excels in computational efficiency. Unlike methods such as MATES\cite{yu2024mates},Group-MATES\cite{yu2025data} and REGMIX, which require training small proxy models, the Group Influence calculation and optimization process of TiKMiX is highly efficient.
In our 1B model experiments, the total computational overhead for \textbf{TiKMiX-D} to determine the next-stage mixture (including influence calculation and regression model inference) was only about \textbf{20\%} of that required by the \textbf{RegMix} method, while achieving comparable or even superior performance. This high efficiency makes TiKMiX a practical and powerful tool for large scale LLM training.

\begin{table}[h]
\centering
\resizebox{\linewidth}{!}{
\begin{tabular}{lcccc}
\toprule

\multirow{2}{*}{\textbf{Benchmark}} & \multicolumn{2}{c}{\textbf{Loss}} & \multicolumn{2}{c}{\textbf{TiKMiX-D}} \\

\cmidrule(lr){2-3} \cmidrule(lr){4-5}

 & \textbf{5B} & \textbf{10B} & \textbf{0.1B} & \textbf{0.5B} \\
 
\midrule
\multicolumn{5}{l}{\textit{\textbf{In-Domain Benchmarks}}} \\
MMLU  \cite{hendrycks2020measuring}         & 31.4 & 31.2 & 32.2 & 32.1 \\
HellaSwag  \cite{zellers2019hellaswag}     & 56.3 & 56.4 & 57.4 & 57.6 \\
ARC Easy  \cite{clark2018think}            & 67.3 & 65.6 & 69.3 & 69.1 \\
ARC Challenge  \cite{clark2018think}       & 34.4 & 33.4 & 37.0 & 37.1 \\
TriviaQA  \cite{joshi2017triviaqa}         & 16.5 & 16.9 & 17.7 & 17.9  \\
\hline
\multicolumn{5}{l}{\textit{\textbf{Out-of-Domain Benchmarks}}} \\
PiQA  \cite{bisk2020piqa}                  & 73.2 & 73.5 & 74.1 & 74.2 \\
OpenBookQA  \cite{mihaylov2018can}         & 36.4 & 36.6 & 37.4 & 37.3 \\
BoolQ  \cite{clark2019boolq}               & 59.4 & 59.7 & 61.3 & 61.5 \\
MathQA  \cite{amini2019mathqa}             & 23.9 & 23.7 & 23.5 & 23.6 \\
\midrule
Average Perf.                            & 44.3 & 44.1 & 45.5 & 45.6 \\
\bottomrule
\end{tabular}
}
\caption{Ablation study of Loss and TiKMiX on different data sizes.}
\label{tab:ablation-study}
\vspace{-1.5em}
\end{table}

\subsection{Ablation Study}

We conduct a series of ablation studies, with the results presented in Table~\ref{tab:ablation-study}. Our primary investigation focused on the efficacy of using \textbf{group influence} and \textbf{TiKMiX} for preference observation and data mixture adjustments. As shown in Table~\ref{tab:ablation-study},  our approach allows for the accurate observation of model preferences using only 0.1B tokens and requires no model training, leading to a significant performance improvement over the loss. This highlights the superiority of our method in efficiently identifying and correcting data biases. We further discuss the effectiveness of our model on a larger scale in the appendix.
\section{Conclusion and Discussions}
\label{sec:Conclusions}

In this work, we address the suboptimality of static data mixing strategies in language model pre-training, demonstrating that a model's learning preferences for different data domains evolve dynamically with its training progress. To tackle this, we introduce TiKMiX, a novel framework that dynamically adjusts the data mixture based on Group Influence, a highly efficient metric to evaluate the contribution of data domains to the model's performance. By framing data mixing as an influence-maximization problem, we developed two approaches: TiKMiX-D, which directly optimizes the mixture and surpasses state-of-the-art methods like REGMIX using only 20\% of the computational resources, and TiKMiX-M, which uses a regression model to predict superior mixtures, achieving an average performance gain of 2\% across 9 downstream benchmarks. Our experiments confirm that dynamically adjusting the data mixture based on Group Influence significantly improves performance by mitigating the "under-digestion" of data seen with static ratios. We plan to conduct further experiments on larger-scale models and more diverse datasets to further validate the effectiveness of Group Influence and TiKMiX.

\bibliography{my}

\newpage
\clearpage 

\section{Appendix}
\label{sec:appendix}

\subsection{Experimental Setup}

\paragraph{Datasets and Models}

Web data serves as one of the core sources for pre-training large language models (LLMs), playing a crucial role in enhancing model capabilities due to its broad coverage and diversity. However, precisely because web data encompasses a wide range of domains—including news, encyclopedias, forums, and academic content—its highly diverse origins make it extremely challenging to achieve a balanced mixture across different domains. We follow the same experimental setup as prior studies on web data mixture~\cite{wettig2025organize,liu2025quadmix}, utilize the RefinedWeb dataset~\cite{penedo2023refinedweb}, and employ the domain classifier~\cite{he2023debertav3} to categorize the data into 26 distinct domains. Our models, ranging in size from 1B to 7B parameters, are trained on up to 1 trillion tokens. The training process is divided into two distinct stages, each consisting of 500 billion tokens, with a strategic adjustment of the data mixture ratio at the transition point between stages. 
We compare TiKMiX against several representative data mixing strategies:
\textbf{Pile-CC \cite{gao2020pile}}: The original data mixture proposed by the authors of The Pile based on heuristics.
\textbf{REGMIX \cite{liu2024regmix}}: SOTA method that uses a regression model to predict and optimize validation loss for determining the mixture.
\textbf{DoReMi \cite{xie2023doremi}}: a classic dynamic data mixing method that relies on a proxy model. \textbf{QUAD \cite{zhangharnessing}}: a method for dynamic selection during training after clustering data We use the best-reported mixture from their paper, re-normalized to the domains available in our setup. 

Our proposed TiKMiX method achieves a balance between dynamic adaptability and computational efficiency in data mixture strategies. Similar to other dynamic approaches such as DoReMi and QUAD, TiKMiX adjusts the data mixture ratios according to the current state of the model. However, unlike these methods, TiKMiX does not require multiple iterations, which significantly improves training efficiency. Furthermore, TiKMiX simplifies the data mixing process and reduces engineering complexity without sacrificing model performance.

To systematically evaluate the effectiveness of different data mixing strategies, we conduct large-scale experiments on the RefinedWeb dataset. Our models range in size from 1B to 7B parameters and are trained on up to 1 trillion tokens. The training process is divided into two distinct stages, each consisting of 500 billion tokens. At the transition point between these two stages, we strategically adjust the data mixture ratios to further assess the impact of mixing strategies on model performance.

\subsection{Downstream Task Evaluation}

To conduct a comprehensive and rigorous evaluation of our proposed method, we curated a diverse suite of nine widely-recognized downstream benchmarks. This evaluation matrix is strategically divided into two categories: \textbf{in-domain} and \textbf{out-of-domain}. This bifurcation allows for a dual-faceted assessment of our model's capabilities: on one hand, to measure its proficiency on tasks closely aligned with its training objectives, and on the other, to critically examine its ability to generalize learned skills to novel tasks and knowledge domains. The consistent performance gains observed across both categories underscore our method's ability to enhance the model's foundational capabilities and foster robust generalization.

\paragraph{In-Domain Evaluation}
Our in-domain evaluation suite is designed to probe the model's core competencies in complex reasoning, commonsense understanding, and knowledge-intensive applications. These benchmarks are thematically aligned with our method's primary optimization goals and serve to quantify the depth of improvement in these critical areas.

\begin{itemize}
    \item \textbf{MMLU (Massive Multitask Language Understanding)}~\cite{hendrycks2020measuring}: A highly challenging multitask benchmark that assesses knowledge across 57 disparate subjects, ranging from elementary mathematics and U.S. history to computer science and law. MMLU demands not only a vast repository of knowledge but also the ability to perform precise, domain-specific reasoning, making it a key indicator of a model's comprehensive intellectual and academic capabilities.

    \item \textbf{HellaSwag}~\cite{zellers2019hellaswag}: A commonsense reasoning benchmark that tasks the model with selecting the most plausible continuation for a given context. HellaSwag is distinguished by its use of adversarially-generated distractors, which are designed to be highly confusable for models that rely on superficial statistical cues. It therefore serves as a robust test of a model's deeper understanding of causality and everyday situations.

    \item \textbf{ARC (AI2 Reasoning Challenge)}~\cite{clark2018think}: This benchmark evaluates reasoning and comprehension on grade-school science questions. We assess performance on both its subsets: \textbf{ARC-Easy (ARC-E)}, which contains questions often solvable via information retrieval, and the more difficult \textbf{ARC-Challenge (ARC-C)}, which requires multi-step reasoning and synthesis of knowledge. Evaluating on both allows for a fine-grained analysis of the model's capabilities, from basic knowledge retrieval to complex scientific inference.

    \item \textbf{TriviaQA}~\cite{joshi2017triviaqa}: A large-scale reading comprehension benchmark where questions are authored by trivia enthusiasts, leading to a high degree of diversity and complexity. The task requires models to locate answers within lengthy, evidence-rich documents, often amidst significant distractor information. It primarily evaluates the model's proficiency in long-context processing, precise information retrieval, and fact verification.
\end{itemize}

\paragraph{Out-of-Domain Evaluation}
To rigorously assess the generalization power of our method, we selected a set of out-of-domain benchmarks that are distinct from the in-domain tasks in terms of subject matter, format, or required reasoning skills. Performance on these benchmarks directly reflects the model's ability to transfer its learned meta-skills to new and unseen challenges.

\begin{itemize}
    \item \textbf{PiQA (Physical Interaction QA)}~\cite{bisk2020piqa}: A commonsense benchmark focused on physical reasoning. Presented in a question-answering format, it requires the model to understand the properties and affordances of everyday objects (e.g., "How can you cool a cup of water faster?"). PiQA probes the model's intuitive grasp of how the physical world operates, a domain of commonsense distinct from academic knowledge, making it an excellent test of generalization.

    \item \textbf{OpenBookQA}~\cite{mihaylov2018can}: This benchmark simulates an "open-book" exam, requiring the model to answer questions using a given set of elementary science facts. Success demands not only reading comprehension but, more importantly, the ability to reason over and combine these facts to answer questions whose solutions are not explicitly stated. It critically evaluates the model's capacity for multi-step reasoning and knowledge application within a constrained context.

    \item \textbf{BoolQ (Boolean Questions)}~\cite{clark2019boolq}: A dataset of naturally occurring yes/no questions, sourced from real user search queries. The challenge lies in the fact that the relationship between the question and the provided evidence passage is often implicit, requiring sophisticated syntactic and semantic analysis to arrive at a correct Boolean judgment. BoolQ effectively measures the model's fine-grained comprehension of natural, conversational language.

    \item \textbf{MathQA}~\cite{amini2019mathqa}: A mathematical reasoning benchmark featuring multi-step word problems. The task requires models to parse natural language descriptions, formulate a correct sequence of operations, and execute them to find a solution. Covering a diverse range of mathematical reasoning categories, MathQA is a crucial benchmark for evaluating a model's symbolic reasoning and logical chain-of-thought capabilities, representing a significant test of higher-order cognitive skills.
\end{itemize}

By systematically evaluating our method across this dual-category, nine-benchmark matrix, we demonstrate that our approach not only enhances performance in core competency areas (as shown by MMLU and ARC-C) but also significantly improves the transfer of these abilities to novel contexts (as evidenced by PiQA and MathQA). This comprehensive improvement across both in-domain and out-of-domain tasks provides strong evidence for the effectiveness and generalizability of our method.

\begin{table}[t!]
\centering
\caption{Ablation study of REGMIX and TiKMiX on 1B and 7B models.}
\resizebox{\columnwidth}{!}{
\begin{tabular}{lcccc}
\toprule
& \multicolumn{2}{c}{\textbf{1B Model}} & \multicolumn{2}{c}{\textbf{7B Model}} \\
\cmidrule(lr){2-3} \cmidrule(lr){4-5}
\textbf{Benchmark} & \textbf{REGMIX} & \textbf{TiKMiX-D} & \textbf{REGMIX} & \textbf{TiKMiX-D} \\
\midrule
\multicolumn{5}{l}{\textit{\textbf{\quad In-Domain Benchmarks}}} \\
MMLU \cite{hendrycks2020measuring}         & 31.5 & 32.2 & 40.7 & 41.5 \\
HellaSwag \cite{zellers2019hellaswag}     & 56.0 & 57.4 & 76.6 & 76.4 \\
ARC Easy \cite{clark2018think}            & 66.2 & 69.3 & 78.5 & 78.4 \\
ARC Challenge \cite{clark2018think}       & 32.2 & 37.0 & 49.4 & 50.2 \\
TriviaQA \cite{joshi2017triviaqa}         & 15.8 & 17.7 & 46.4 & 45.3 \\
\cmidrule(lr){1-5}
\multicolumn{5}{l}{\textit{\textbf{\quad Out-of-Domain Benchmarks}}} \\
PiQA \cite{bisk2020piqa}                  & 73.3 & 74.1 & 79.1 & 79.2 \\
OpenBookQA \cite{mihaylov2018can}         & 37.0 & 37.4 & 43.2 & 45.4 \\
MathQA \cite{amini2019mathqa}             & 23.2 & 23.5 & 28.8 & 29.9 \\
\midrule
Average Perf.                            & 43.9 & 45.5 & 55.3 & 56.0 \\
\bottomrule
\end{tabular}
}
\label{tab:ablation-study-with-citations}
\end{table}

\begin{figure*}[!t] 
    \centering
    \includegraphics[width=\textwidth]{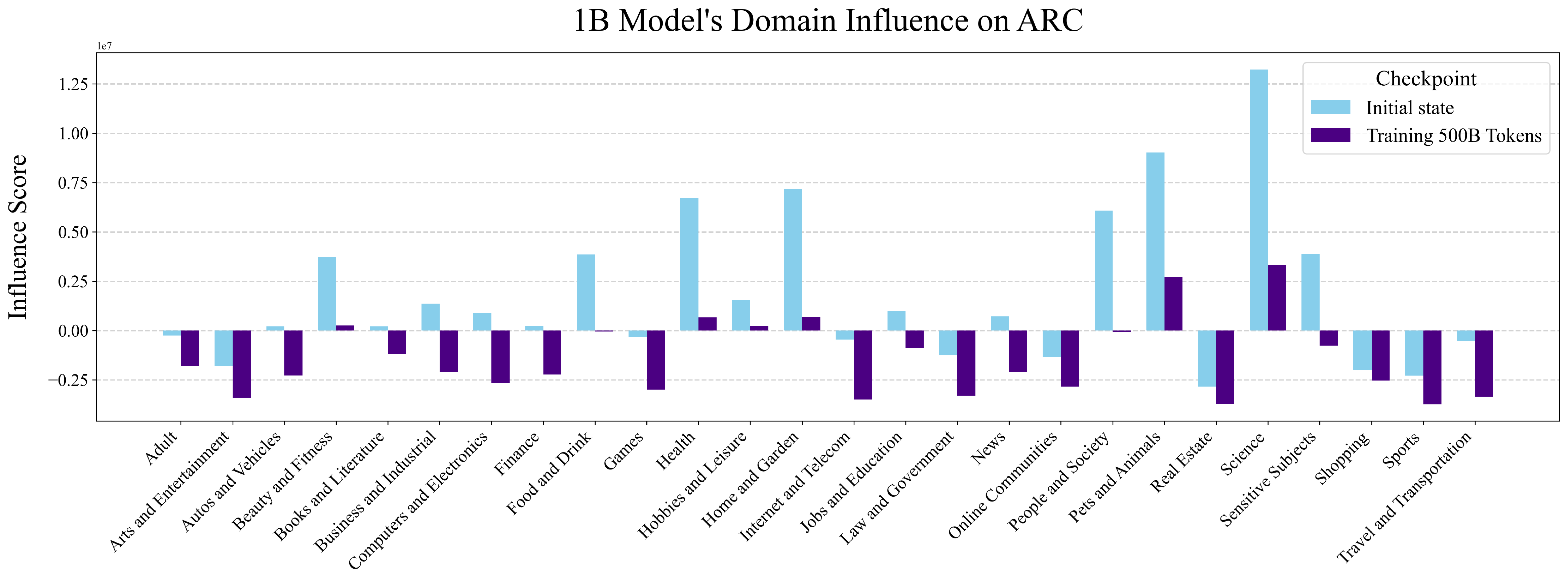}
    \caption{The impact of domains on a 1B model's performance on the ARC benchmark as training progresses.}
    \label{fig:case1}
\end{figure*}

\begin{figure*}[!h] 
    \centering
    \includegraphics[width=\textwidth]{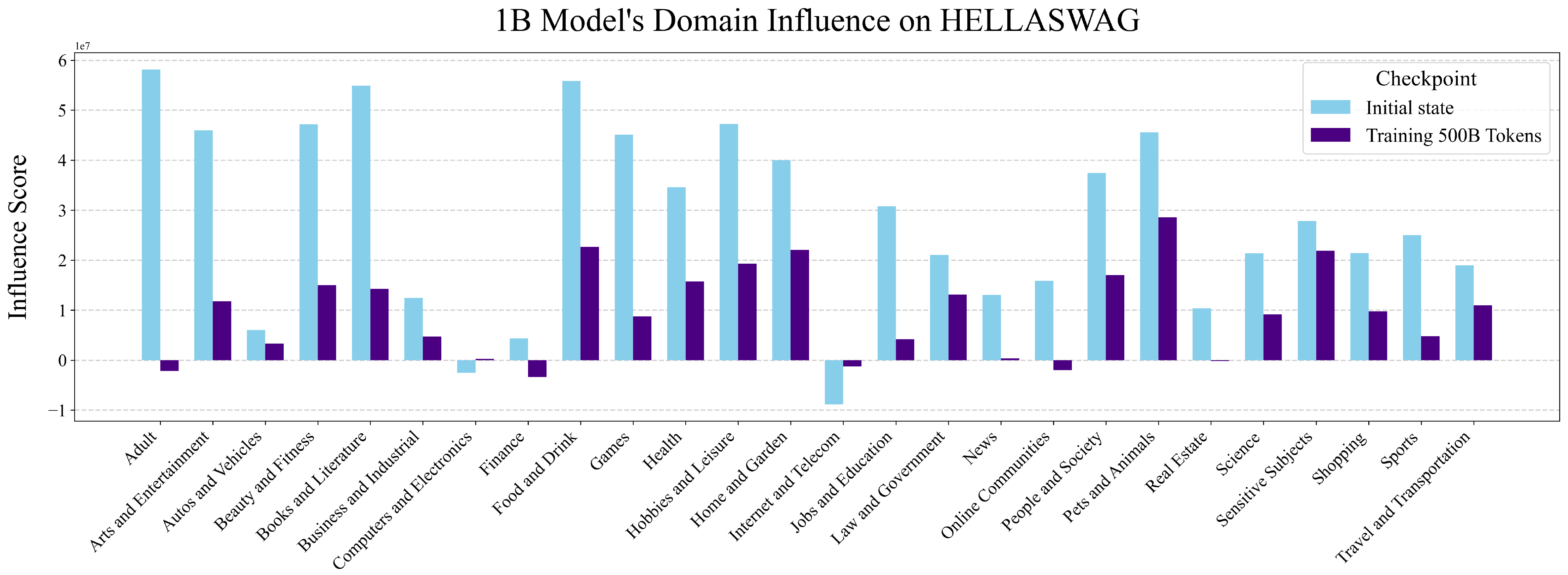}
    \caption{The impact of domains on a 1B model's performance on the HELLASWAG benchmark as training progresses.}
    \label{fig:case2}
\end{figure*}

\begin{figure*}[!t] 
    \centering
    \includegraphics[width=\textwidth]{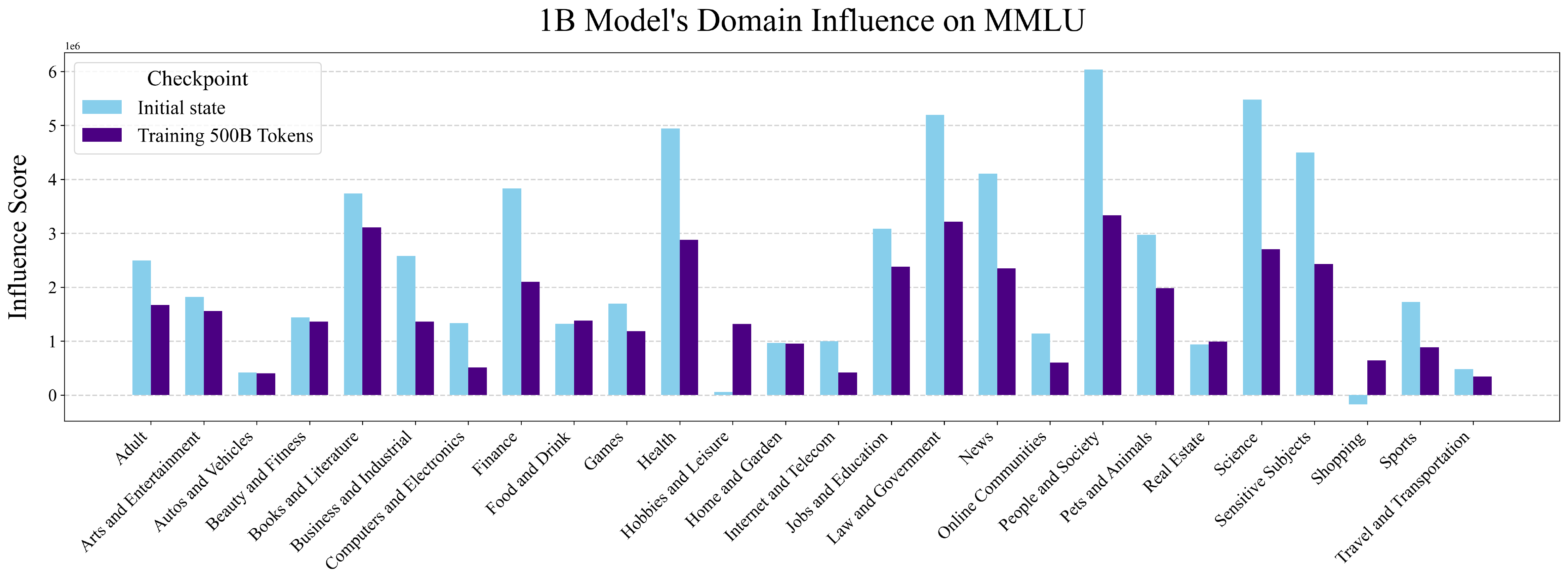}
    \caption{The impact of domains on a 1B model's performance on the MMLU benchmark as training progresses.}
    \label{fig:case3}
\end{figure*}

\begin{figure*}[!t]
    \centering
    \includegraphics[width=\textwidth]{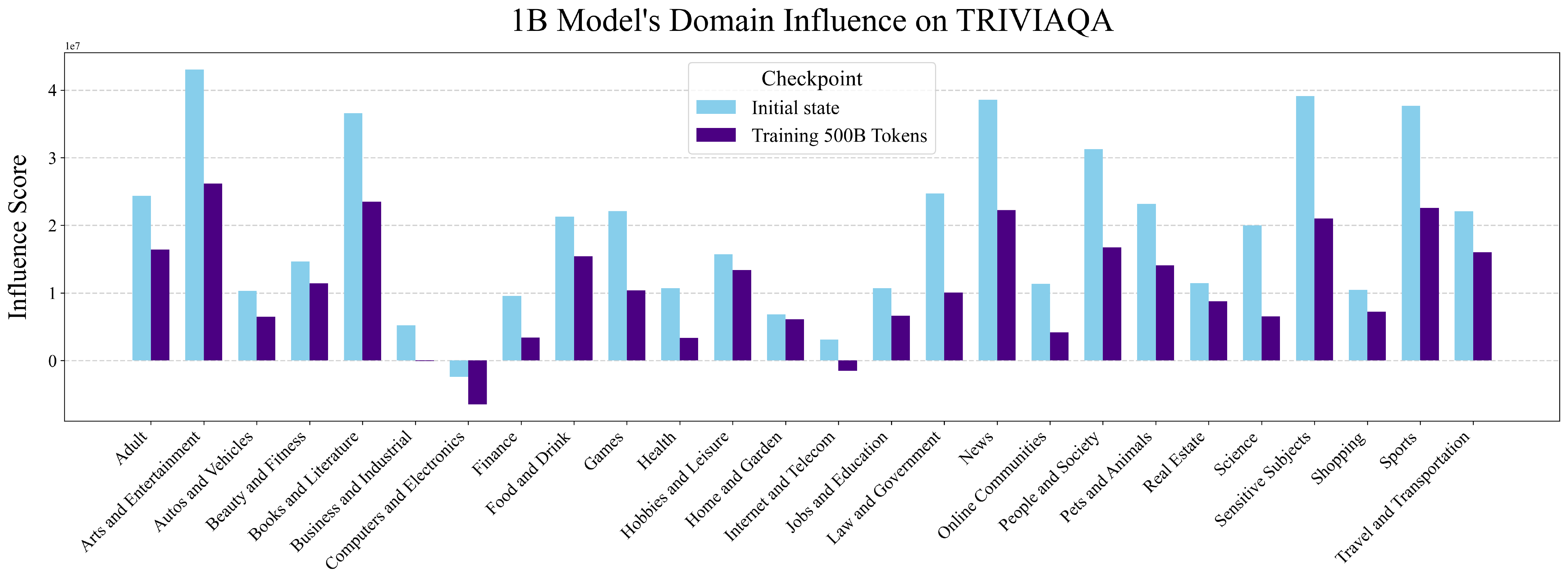}
    \caption{The impact of domains on a 1B model's performance on the TRIVIAQA benchmark as training progresses.}
    \label{fig:case4}
\end{figure*}

\begin{figure*}[!t] 
    \centering
    \includegraphics[width=\textwidth]{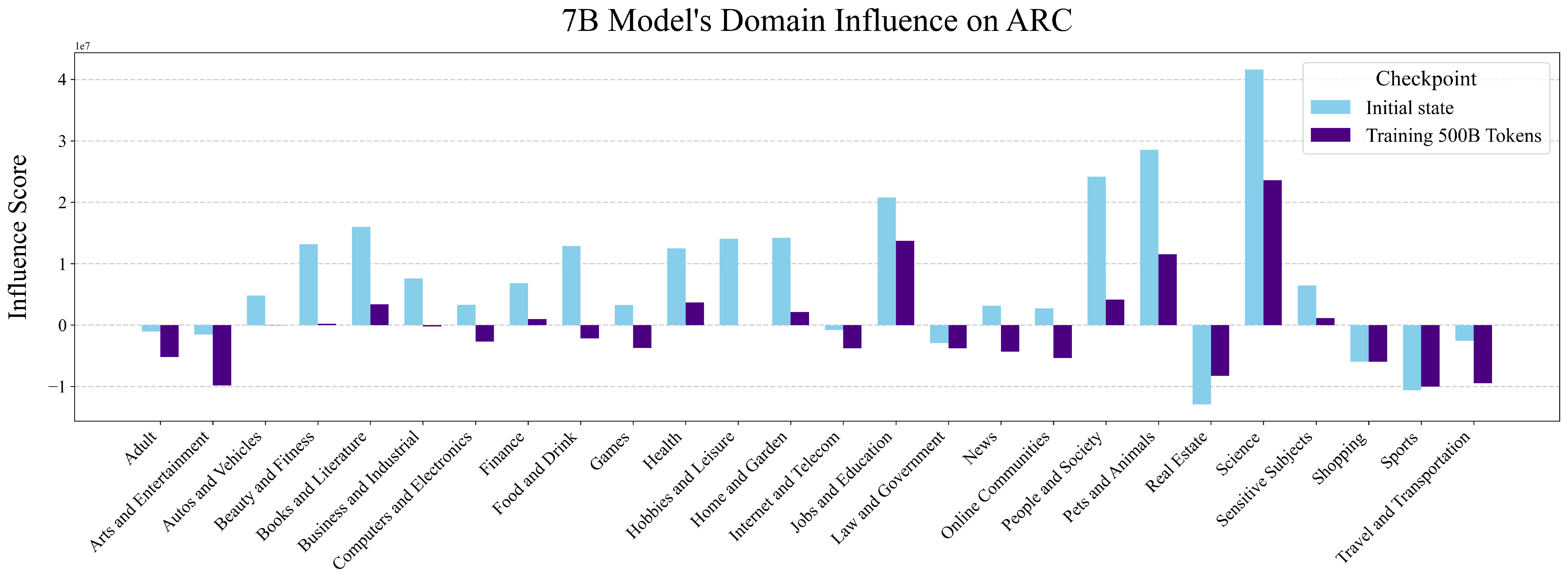}
    \caption{The impact of domains on a 7B model's performance on the ARC benchmark as training progresses.}
    \label{fig:case5}
\end{figure*}

\begin{figure*}[!t] 
    \centering
    \includegraphics[width=\textwidth]{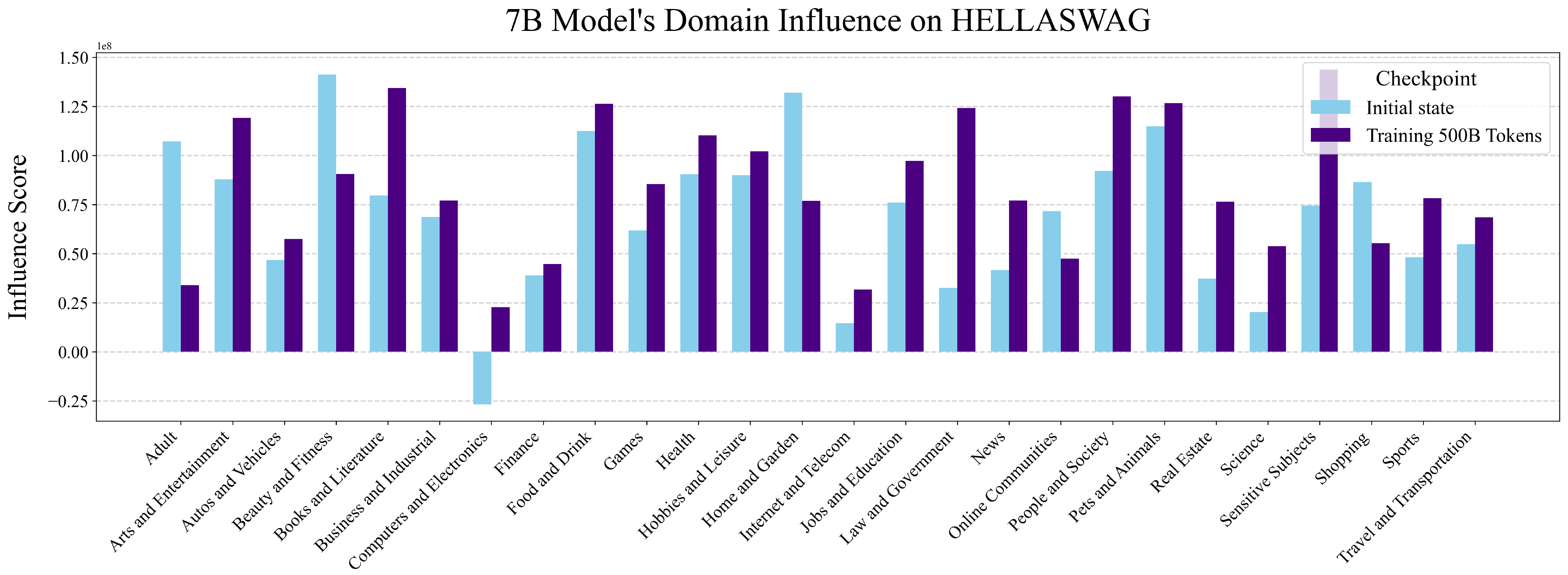}
    \caption{The impact of domains on a 7B model's performance on the HELLASWAG benchmark as training progresses.}
    \label{fig:case6}
\end{figure*}

\begin{figure*}[!t] 
    \centering
    \includegraphics[width=\textwidth]{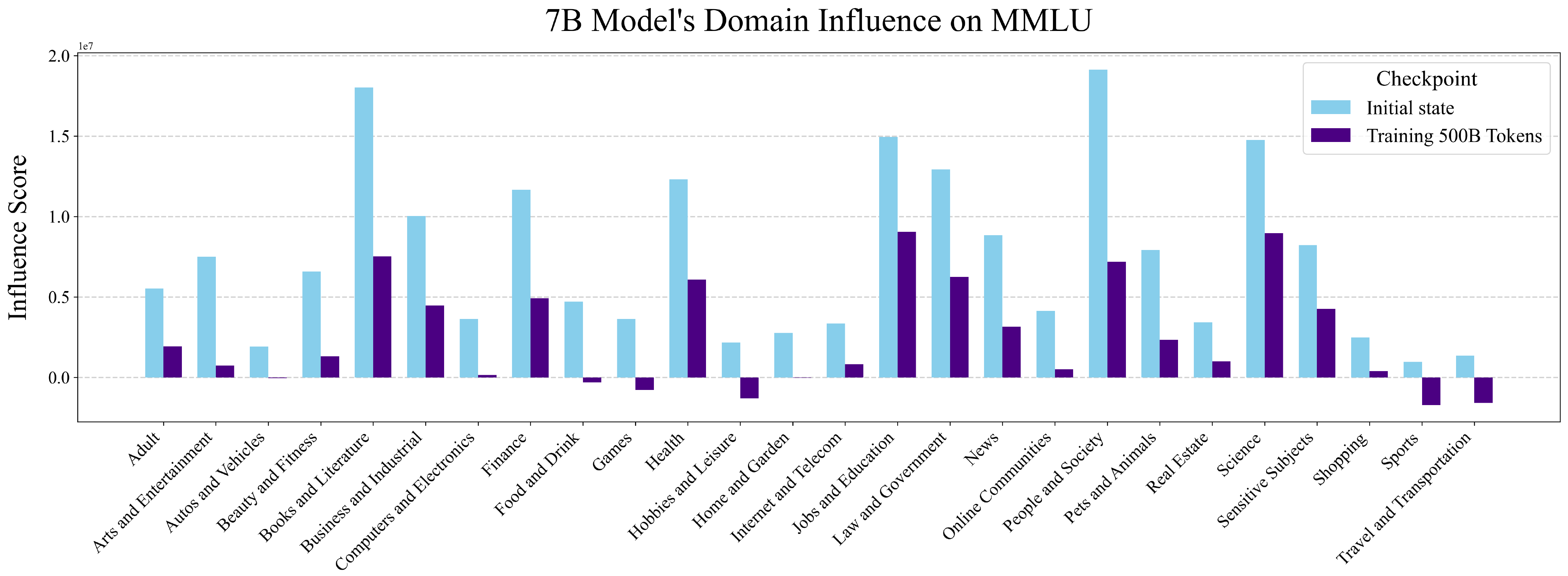}
    \caption{The impact of domains on a 7B model's performance on the MMLU benchmark as training progresses.}
    \label{fig:case8}
\end{figure*}

To further investigate the impact of model scale on data utilization, we present a supplementary analysis in Figures~\ref{fig:case1} to~\ref{fig:case8}. 
Our key finding is that models of different scales (1B and 7B) exhibit significantly different learning responses and form distinct preferences, even when trained on the exact same data. 
This phenomenon reveals a complex interplay between data utility and model scale. 
It provides a solid theoretical foundation for understanding and optimizing the data mixture for models of varying sizes.

\subsection{Experiments on models of different sizes}
Considering computational overhead, for the 7-byte model, we adopted an experimental design similar to REGMIX\cite{liu2024regmix}, training with 500B tokens in the first stage and 200B tokens in the second stage.
Table~\ref{tab:ablation-study-with-citations} presents the experimental results of our method on models of different scales.
It can be observed that our proposed method significantly outperforms the current state-of-the-art approach, REGMIX, on both in-domain and out-of-domain benchmarks.
The performance on the 7B model effectively demonstrates the scalability of our approach.
Furthermore, we note that unlike the 1B model, the 7B model's performance on the benchmarks consistently improves throughout the training process.
This suggests that the advantage of TiKMiX could be even more pronounced with additional training data.

\subsection{Observation of data mixing with Group Influence}

\begin{figure*}[!h] 
    \centering
    \includegraphics[width=\textwidth]{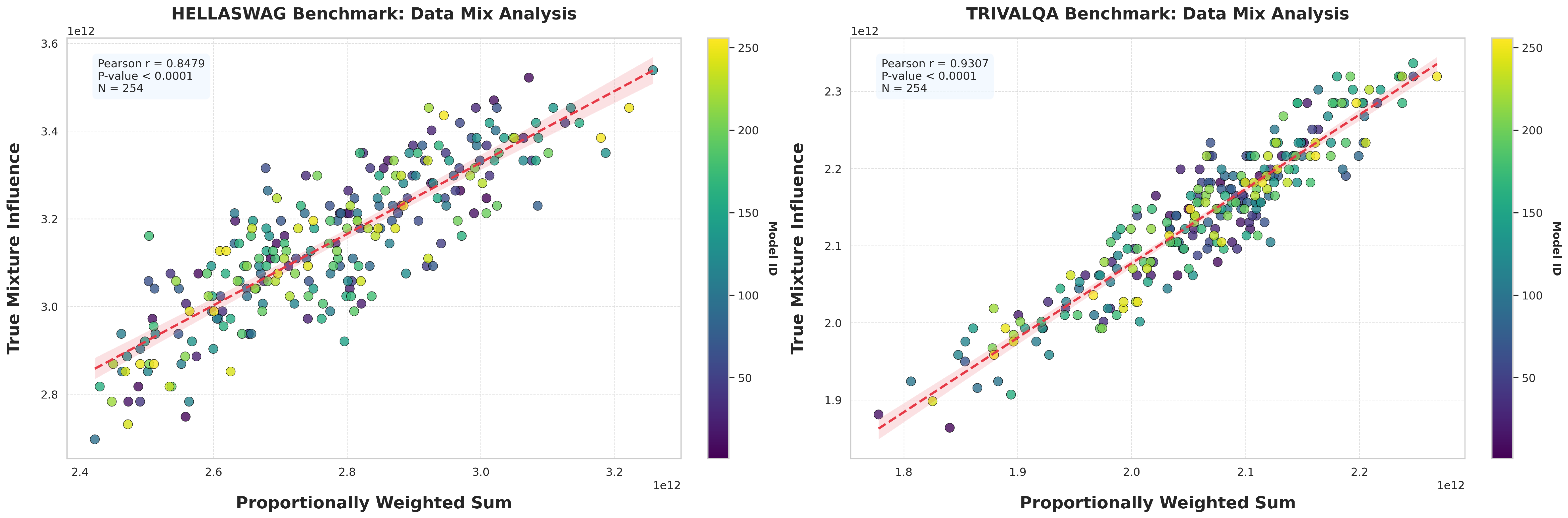}
    \caption{A Group Influence-based Analysis of Data Mixing Effects on Various Benchmarks.}
    \label{fig:case9}
\end{figure*}

To conduct a rigorous analysis of inter-domain interactions during mixed training, we designed an experiment to test the principle of influence additivity. Our hypothesis was that the influence of a mixed dataset on a validation set could be accurately predicted by a weighted sum of the influences from its individual constituent domains. To verify this, we first established a baseline mixing recipe using our TiKMiX-D method. We then systematically explored the local space around this recipe by generating 256 perturbed configurations, created by applying a random scaling factor between 0.5 and 2.0 to each domain's original proportion. After filtering out two sampling outliers, we proceeded with 254 unique data mixture configurations. For each of these 254 points, we sampled a corresponding 0.1B token dataset and measured its direct influence. We then compared this empirical influence value against a predicted influence, which was calculated by summing the pre-computed influences of each individual domain, weighted by their respective proportions in the mixture. As depicted in Fig~\ref{fig:case9} , this comparison revealed a strong linear correlation. Specifically, the Pearson correlation coefficients on the ARC\cite{clark2018think}, Hellaswag\cite{zellers2019hellaswag}, and TriviaQA\cite{joshi2017triviaqa} benchmarks reached 0.845, 0.848, and 0.931, respectively, all of which are statistically highly significant ($p < 0.0001$). This result provides compelling evidence that the outcome of data mixing is highly predictable and can be modeled as a linear combination of inter-domain influences. Consequently, this finding offers a solid empirical justification for the theoretical soundness of our proposed two-stage optimization framework, encompassing both TiKMiX-D and TiKMiX-M.

\clearpage

\begin{figure*}[!t] 
    \centering
    \includegraphics[width=\textwidth]{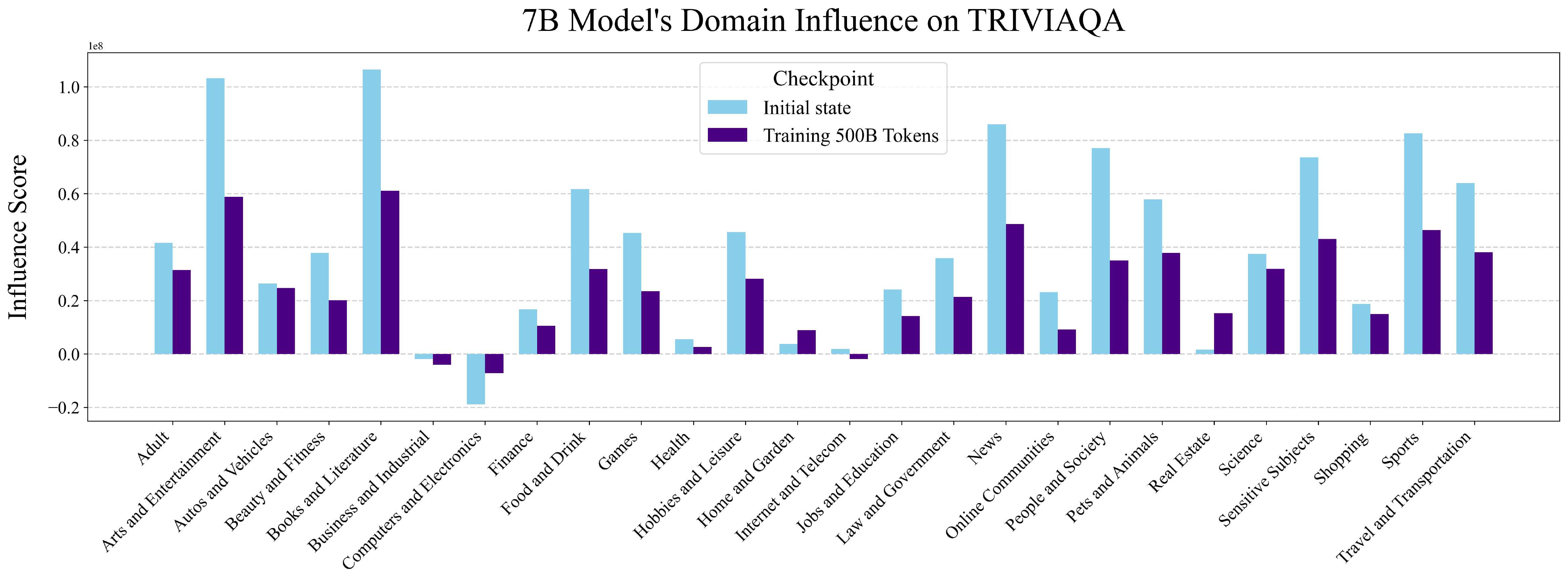}
    \caption{The impact of domains on a 7B model's performance on the TRIVIAQA benchmark as training progresses.}
    \label{fig:case7}
\end{figure*}

\clearpage

\end{document}